**An initialization method for the *k*-means using the concept of useful nearest centers**

Hassan Ismkhan[1]

**Abstract**

The aim of the *k*-means is to minimize squared sum of Euclidean distance from the mean (SSEDM) of each cluster. The *k*-means can effectively optimize this function, but it is too sensitive for initial centers (seeds). This paper proposed a method for initialization of the *k*-means using the concept of useful nearest center for each data point.

## 1. Introduction

Let $I = \{x_1, x_2 \ldots x_n\}$ be an instance with *n* number of data points, and each data point is characterized by *m* variables: $x_i = \{x_{i,1}, x_{i,2}, \ldots x_{i,m}\}$. The aim of the *k*-means [1] is to group these data points into the *k* disjoint subsets $S_i$ ($0 \leq i \leq k$), such that minimize the sum of squared Euclidean distances to the mean of each subset (SSEDM) as its objective function. For number *k*, input instance *I*, and partitioning solution *S*, where $S = S_1 \cup S_2 \cup \ldots \cup S_k$, SSEDM can be defined as following equation:

$$SSEDM(S) = SSEDM(I, S, k) = \sum_{i=1}^{k} SSEDM(S_i) \qquad (1)$$

Where *SSEDM(S_i)* is partial SSEDM of cluster $S_i$ and can be defined as following equation:

$$SSEDM(S_i) = \sum_{\forall P \in S_i} dis(P, \ mean(S_i))^2 \qquad (2)$$

[1] Department of Computer Engineering, University of Bonab.
 Email: h.ismkhan@bonabu.ac.ir, esmkhan@gmail.com

Here, *dis(,)* indicates Euclidian distance between two data points, and *mean($S_i$)* indicates center of cluster $S_i$.

Although exactly minimizing SSEDM can not guarantee the best quality for clustering solution [2], and exactly minimizing this function is an *NP*-Hard problem [3], not only this function is popular to measure quality of clustering solutions, it is also used as an objective function in the *k*-means, which is one of the most influential data mining algorithms [4]. The *k*-means operates instructions as follows: (1) select *k* number of points as initial centers, (2) assign each data point to its nearest center, (3) updates location of centers, (4) repeat instructions 2 and 3 until convergence. The *k*-means can effectively minimize SSEDM for instances with linearly separable clusters [18], but it is too sensitive to initial centers. Therefore, many initialization methods are proposed in the literature. References [5] [6] propose an initialization method, namely Maxmin, which chooses the first center randomly, and in the rest, it selects a data point with largest distance to its nearest center as a new center.

Among famous initialization methods, k-means++ [7] is one of the most accurate algorithm and easy to be implemented. k-means++ chooses the first center, among data points of data set, randomly. It chooses a data point $x \in X$ as the *i*-th center with probability $\frac{D(X)^2}{\sum_{x \in X} D(X)^2}$, where *D(x)* denotes the shortest distance from a data point to the closest center we have already chosen.

As the main aim of the *k*-means is to minimize SSEDM, this paper proposes a novel method. It is based on the concept of useful centers of each data point. Results of performed experiments shows how it can outperform k-means++ on real world datasets.

## 2. The proposed initialization algorithm

Before stating the proposed initialization method, we need to state two simple definitions.

Definition 1. The concept of useless nearest center: a center $C$ is useless for a data point $P$, if for a center $C_x$: $dis(P, C_x) < dis(P, C)$ and $dis(C, C_x) < dis(P, C)$.

Definition 2. The concept of useful nearest center: a center $C$ is useful for a data point $P$, if it is not a useless nearest center of $P$.

Example 3.1. In Figure. 1, $C_1$ and $C_2$ are useful center for $P$. $C_3$ is closer to $P$ than $C_1$, but $C_3$ is not a useful center for $P$, because $C_2$ causes that definition 2 does not meet for $P$ and $C_3$.

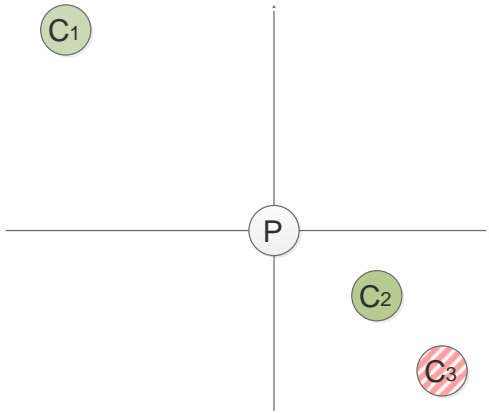

Figure. 1. $C_1$ and $C_2$ are useful center for $P$, but $C_3$ is not a useful center for $P$.

The proposed initialization method, selects the first center with the smallest value on the first axis, then in each of next iterations, a data point $P$ with the largest value of $\frac{\text{average}_{\forall c \in UNC_P}(dis(P,c))}{\max_{\forall c \in UNC_P}(dis(P,c))} \times \sum_{\forall c \in UNC_P} \ln(dis(P,c))$ is selected as the next center, where $UNC_P$ is the set of useful nearest centers of $P$. This process is continued until required number of centers are selected. It should be noted that in implementation, if a recently added center $C_{i-1}$ is useless for $P$, then it is ignored; otherwise, it is added to $UNC_P$, and in this case, some data points may be dropped out from $UNC_P$.

## 3. Results of experiments

To evaluate performance of the proposed initialization algorithm, here, call UNC-KM, in terms of SSEDM, it competes with the *k*-means (KM) and k-means++ (KM++) on real world datasets including iris, human activity recognition (HAR), and shuttle which are available via UCI machine learning repository. The results are introduced in Table. 1. For iris and shuttle, UNC-KM wins with high gap of obtained SSEDM.

| | Table. 1. Result of competitors on UCI datasets | | | | | |
|---|---|---|---|---|---|---|
| | **Average SSEDM of 50 runs** | | | **Average runtime (second) of 50 runs** | | |
| | Iris | HAR | shuttle | Iris | HAR | shuttle |
| **KM** | 9.95E+01 | 184867.5 | 8.13E+08 | 0.00E+00 | 1.70E+00 | 4.26E-01 |
| **MK++** | 8.41E+01 | 184820.3 | 7.99E+08 | 0.00E+00 | 1.64E+00 | 4.34E-01 |
| **UNC-KM** | 7.89E+01 | 204329.6 | 5.44E+08 | 0.00E+00 | 2.80E+00 | 2.99E-01 |

**Conclusion**

This paper propose a novel and simple method to initialize centers of the *k*-means. The proposed method is based on the concept of useful nearest center which is defined for each data point. After each iteration, when a new center is added, the list of useful nearest center of each data point is updated. During each iteration, a data point with the largest amount of a value, which is depends on distance from its nearest center, is selected as the next center. In comparison to the random *k*-means and k-means++, not only the proposed algorithm has acceptable runtime, it obtain better SSEDM with high gap from the random *k*-means and k-means++.

**Acknowledgment**

This research is supported by University of Bonab, via author personal grant.